\documentclass[letterpaper, 10 pt, conference]{ieeeconf}  

\IEEEoverridecommandlockouts                              

\usepackage{graphics} 
\usepackage{epsfig} 
\usepackage{mathptmx} 
\usepackage{times} 
\usepackage{amsmath} 
\usepackage{amssymb}  
\usepackage{algorithm}
\usepackage{epstopdf}
\let\labelindent\relax
\usepackage{enumitem}
\usepackage[table,xcdraw]{xcolor}
\usepackage{hyperref}
\title{\LARGE \bf
Concurrent Policy Blending and System Identification for Generalized Assistive Control
}

\author{Luke Bhan$^{1}$, Marcos Quinones-Grueiro$^{1}$, and Gautam Biswas$^{1}$
  \thanks{$^{1}$ All authors are currently members of Vanderbilt University, Nashville, TN. Correspondence to luke.bhan@vanderbilt.edu}%
}

\begin{document}

\maketitle
\thispagestyle{empty}
\pagestyle{empty}

\begin{abstract}
  In this work, we address the problem of solving complex collaborative robotic tasks subject to multiple varying parameters. Our approach combines simultaneous policy blending with system identification to create generalized policies that are robust to changes in system parameters. We employ a blending network whose state space relies solely on parameter estimates from a system identification technique. As a result, this blending network learns how to handle parameter changes instead of trying to learn how to solve the task for a generalized parameter set simultaneously. We demonstrate our scheme's ability on a collaborative robot and human itching task in which the human has motor impairments. We then showcase our approach's efficiency with a variety of system identification techniques when compared to standard domain randomization. The code is available on \href{https://github.com/lukebhan/assistive-gym}{Luke Bhan's Github}.
\end{abstract}

\section{INTRODUCTION}
Over the last few years, there has been significant interest
in developing models that are trained in simulation and then
transferred to the real world \cite{one}, \cite{two}, \cite{three}. Despite progress in 
learning from simulated policies, these methods still suffer from long simulation times as they require large amounts of experience to handle the unknown environments present in the real world. Additionally, these policies struggle to generalize for complex situations as they may become unpredictable when faced with new challenges. As such, researchers have approached training these policies in two distinct ways. The first approach utilizes techniques, such as Kalman Filters to identify system parameters that can inform policies on how to respond in different environmental conditions \cite{six}. However, these policies struggle to generalize and often require re-tuning \cite{six}. The second approach involves randomizing sets of system parameters during training so that the policy learns to robustly handle a wide variety of situations. This approach - domain randomization \cite{five} - requires the actual parameters to be in the set of random values and as such, the robustness of the policy is directly correlated to the range of the generated randomized parameters. Moreover, the resulting policies are overly conservative in action selection, especially when the parameter set is large. Finally, for complex tasks with many parameters, creating large ranges across multiple parameter values requires significant training time before a robust policy can be generated \cite{four}.

Both of the above-mentioned approaches require a single policy that aims to 1) solve the task at hand for a single set of parameters and 2) can generalize the solution to a wide set of parameters. These policies struggle to generalize and fail for parameter spaces in complex tasks. As an alternative, we attempt to decouple the process of generating generalized system parameters to support the learning of policies that solve the task efficiently. We do this by utilizing blending techniques informed by system identification methods. With our approach, we train single policies that are efficient for a distinct sets of the parameter space and utilize a blending network to identify how best to combine the actions of these individual policies based on the estimated parameters of the system. We verify the validity of our approach for a collaborative robot and human itching task in which the human has different combinations of motor impairments. Our proposed approach works with the assumption that the space of each parameter is bounded, thus any combination of parameter values results in a convex space. This guarantees convergence of the blending network given a set of individual policies learned to control the system under single parameter variability.

Our paper makes the following primary contributions:
\begin{itemize}
  \item We decouple the process of learning a single task and generalizing to a large set of system parameter combinations using a blending network technique; 
   \item We design an architecture, which integrates a blending network that accurately handles the generalization of its sub-models to system parameters; and
   \item We implement our scheme on a simulated collaborative human and robotic locomotion task with various environment parameters to demonstrate its effectiveness across different system identification methods.
 \end{itemize}

\section{Related Work}
There have been many approaches to policy learning based on estimation of simulation parameters; however, to the author's knowledge, none combined system identification with a blending approach. For example, \cite{eight} demonstrates the use of simultaneous system identification and policy training where they explore a series of predictive error methods to minimize the difference between the observed parameters and  the estimated parameters for model predictive control (MPC). Additionally, domain randomization has been used for a motorized robotic control tasks \cite{nine}. However, these tasks do not consider a collaborative environment nor do they handle multiple faults introduced by the interacting agents. Furthermore, \cite{ten} solves a challenging Rubik's cube control task by automatic domain randomization that slowly increases the difficulty of the task, but can take significant time as it does not consider the integration of any real-world sampling. Lastly, \cite{fourteen} considers an adaptive domain randomization strategy where the framework attempts to identify domains that can create challenging environments for the policy. This approach is similar to ours, except that their approach is purely data driven based on the result of their policy. This requires significant training time due to potential sample inefficiency. In contrast, we utilize prior domain knowledge to identify environments that have a high potential of being challenging for the policy. 

In addition to the large amount of research designed for efficient sim2real transfer, there have been a series of recent work that demonstrate the effectiveness of policy blending. \cite{seven} demonstrates the use of policy blending for simple tasks such as opening a cap, flipping a breaker, and turning a dial. However, their policy learns directly from sensor measurements and does not consider impairments in the agents. Furthermore, \cite{eleven} has shown a policy blending technique between a human and robot policy for robot-assisted control to accurately assist the human with various tasks such as fetching a water bottle. However, this work does not consider training models using modern deep reinforcement learning (DRL) techniques. Given these approaches, it is worthwhile to combine robust policy blending with modern system identification as a new approach to generalized modelling.
\section{Background}
In this section, we first formalize the reinforcement learning problem for tasks with multiple varying parameters. Then, we outline three approaches presented in the literature to solve such tasks. 

\subsection{Robust Markov Decision Process}
We model tasks  with  multiple  varying  parameters as a robust Markov decision process (R-MDP) defined using a tuple $(S,A,R, \mathcal{P},\gamma)$, where $S$ is the state space,
$A$ the action space, $R : S \times  A \rightarrow \mathbb{R}$ represents a reward function, $\gamma \in [0,1]$ is called a discount factor, and $\mathcal{P}(s,a) \in \mathcal{M}(S)$ is an uncertainty probability set where $\mathcal{M}(S)$ represents a family of probability measures over next states $s' \in S$. The next state of the system is contingent on the conditional measure $p(s'|s,a) \in \mathcal{P}(s,a)$ where $s \in S$ is the current state and $a \in A$ is the action selected by an agent.

In this work, we adopt the assumption that ${\cal P}$ is structured as a Cartesian product $\otimes_{s\in S, a\in A} {\cal P}_{s,a}$, also known as the state-action rectangularity assumption \cite{wiesemann2013robust}. This implies that  nature can choose the worst-transition independently for each state and action. Moreover, we assume that the uncertainty probability set is defined by the parameter space that characterizes the task $\lambda_t \in \Lambda$ such that  $P(\mathbf{s}_{t+1}=\mathbf{s}'|\mathbf{s}_t,\lambda_t,\mathbf{a}_t,\boldsymbol{\kappa})$ ($ S \times \Lambda \times A  \rightarrow S$).

In a standard MDP framework, a policy $\pi: S \rightarrow p(A)$ maps states to distributions over actions with the goal of maximizing the sum of the discounted rewards over the future. The optimization criterion is the following
\begin{equation}
\label{opt}
\cal J(\pi^*)= \max_{\pi \in \Pi} V^{\pi}(s) \; , \; \forall s \in \mathit{S}.
\end{equation}
where the value function, $V^{\pi}: \mathit{S} \rightarrow \mathit{R}$, defines the value of being in any given state $s$ 
\begin{equation}
\label{vf1}
V^{\pi}(s) = E\left[ \sum_{t=0}^{\infty} \gamma^{t}\mathit{R}(s,a)\right] \; , \; \forall s \in \mathit{S}.
\end{equation} 

Traditional RL algorithms require that the system dynamics and reward function do not change over time to be able to find an optimal deterministic Markovian policy satisfying \ref{opt}. This property is clearly not satisfied in the R-MDP case. Therefore, we propose an approach to decouple learning how the system dynamics change depending on the system parameters and learning how to optimally solve the task. 

\subsection{Domain Randomization}
To effectively identify $\pi^*$ in simulation, a set of parameters must be defined to model the environment. Domain randomization attempts to sample a set of some N parameters which we will denote $\xi$ for which a reasonable range of potential values is constructed - usually from domain specific knowledge \cite{two}. In this paper, we will consider domain randomization of the uniform type such that the parameters are uniformly sampled within a feasible range. For example, the weakness of a certain human joint can be sampled uniformly between 0 and 1 where 0 invokes no mobility while 1 is a joint that is at full strength. 

\subsection{System Identification Via Parameter Estimation}
System identification through parameter estimation is a well studied subject in which a estimator can consistently receive samples from a real world environment and generalize these samples into an estimated true value. In this work, we utilize the Unscented Kalman Filter (UKF) for our estimator \cite{fifteen} and make the assumption that our real-world parameters can be measured with some confidence, but may be cluttered with noise. Although this is a strong assumption, it can be softened for systems with non-measurable parameters by using approximate models of the system plus general estimation methods like Particle Filtering. See for example \cite{SHAMRAO201879}.

\subsection{Autotuned Search Parameter Model (SPM)}
When the environment's parameters cannot be measured neither estimated through physics-based estimation, we propose to utilize a new technique that can estimate the parameters by interacting with the environment as an agent. Recently, Du et al. formulated a new approach to system identification where they define a data-driven model that learns a map from observation-action-parameter estimate sequences to a probability distribution, i.e.  $(o_{1:T}, a_{1:T}, \xi_{guess}) \implies (0, 1)^N$ such that the parameter estimates $\xi_{guess}$ are greater than, less than, or equal to the true parameters \cite{sixteen}. The mapping works like a binary classifier that is continuously trained concurrently to the policy such that it slowly converges to the real world parameters by learning from its own policy interaction trajectories. Following this iterative search, we can then perform a level of system identification that does not completely rely on domain knowledge for our experiments. 

\section{Approach}
\begin{figure}
  \includegraphics[width=0.5\textwidth]{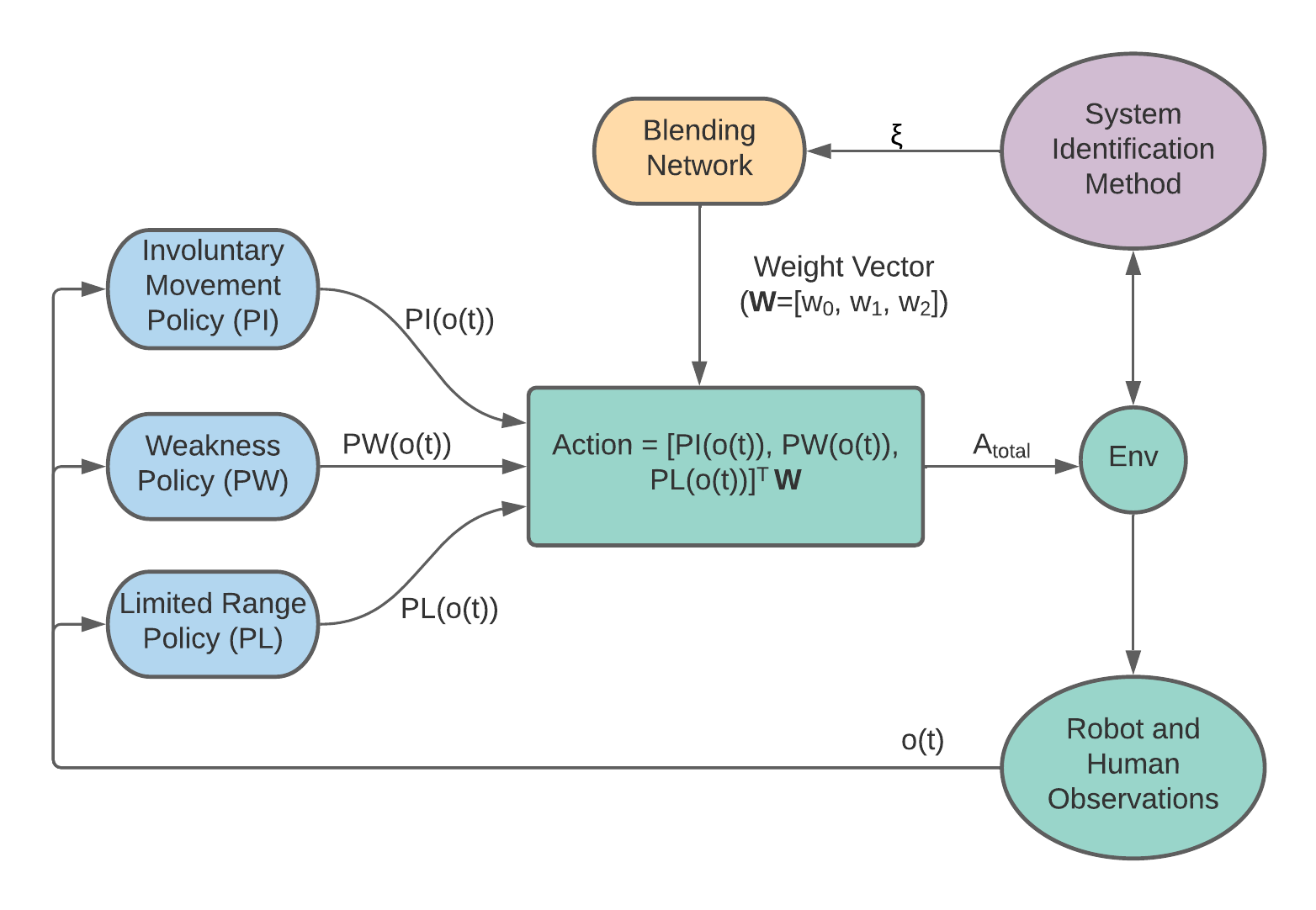}
  \caption{Policy blending with system identification for a human-robot collaborative task}
  \label{fig:arch}
\end{figure}

In this section, we present the details of our solution scheme for solving collaborative tasks with multiple varying parameters. Figure \ref{fig:arch} show the different components of the proposed approach instantiated for the case study presented in the next section. A set of individual policies learned for individual parameter changes is required. Each individual policy can be trained for a single parameter distribution and the total number of policies scales linearly with respect to the total number of parameters. Training each policy can be accomplished through Domain Randomization techniques. The core components of the proposed scheme - the blending network and system identification technique - are described next.

\subsection{Blending Network Learning}
To create a decoupled policy in which we can solve individual tasks while maintaining robustness to a variety of system parameters, we introduce a blending network in which we consider solely the N system parameters $\xi$ as its state space. This policy then only needs to output the weights $\mathbf{W}$ of its sub-policies at each time step to generate the action for the environment. The sub-policies of this model are trained on a single set of constant system parameters in which a unique environment is identified through previous domain knowledge. We then define the action of the blending network as 
$a = \frac{1}{N}\sum_{i=0}^N w_i \pi_i(s_t)$ where $w_i \in \mathbb{R}$, N is the number of sub policies and $\pi_i(s_t)$ represents the action taken by the sub policy given a state at time t.

\subsection{Concurrent System Identification}
We then combine the blending network with a concurrent system identification scheme to obtain a generalized policy that is robust to different environmental changes, i.e. different combinations of parameter values. To do this, we let the state space of the blending network consist of only the estimated parameters and as such must learn to associate certain parameters with the appropriate sub-policies. In practice, every certain number of training steps, we utilize our system identification method to update the state space of the blending network with a more accurate set of estimate system parameters and continue training. We emphasize that our approach is independent of the system identification method chosen and thus can be tailored based on the available domain knowledge of the environment. 

\section{Experiments}

\subsection{Assistive Gym}
For our experiments, we utilize a framework introduced by \cite{seventeen} which is an Open AI gym environment for collaborative human and robot interaction. Assistive gym is a realistic physics environment powered by PyBullet that enforces realistic human joints as well as it provides a series of robots for collective tasks. We utilize the Jaco robot as it performed best in the individual itching policies explored in \cite{seventeen} for our control task and define the success value of a task as the amount of force applied to the target itch position throughout the entire episode. Each episodes consists of 200 time steps equating to 20 seconds of real-world time. 

\subsection{Training of Sub-Policies}
For demonstrating our model, we attempt to solve a collaborative itching task using assistive gym \cite{seventeen} where a robot is assisting an impaired human in itching. We consider 3 impairments similar to \cite{eighteen} for the human:
\begin{enumerate}[label=(\alph*)]
  \item Involuntary Movement: The first impairment is involuntary movement which is handled by adding noise normally distributed to the joint actions of the human. For this policy, we sample the noise according to a normal distribution where each joint in the arm has a mean of 0 noise and a standard deviation of 5 degrees of noise. 
  \item Weakened Strength: The second impairment involves  weakness in the ability for the human to move their arms which is introduced by lowering the strength factor in the PID controller of the joints. This value is also sampled normally with a mean of 0.66 and standard deviation of 0.2 with 1 representing full strength and 0 representing immobility.
  \item Limited Range: Lastly, we consider a limitation in the range of movement for each joint in the arm of the human. Like above, full joint movement is represented by 1 and immobile joints are represented by 0. As such, we sample the limited movement from a normal distribution with mean 0.75 and standard deviation of 0.1. 
\end{enumerate}
Initially, we begin by training a single policy for each individual impairment on 2 million time steps (5000 episodes) using Proximal Policy Optimization (PPO) \cite{nineteen}. We designed a grid-search experiment to obtain the neural network architecture. The best configuration for each network obtained consists of 2 layers of 64 nodes. For all single impairment policies, we define a state space of 64 joints between the robot and human along with an action space of 17 joint targets. All policies use the same reward function as defined in \cite{seventeen}. This reward function considers a weighted combination of the distance of the robot arm to the target itch position, a penalty for large actions, and the contact induced with the itch target: $R(t) = w_{distance}*||pos_{target}-pos_{robot}|| + w_{action}*||a(t)|| + w_{force}*F_{target}$ where $w_i$ represents a weight for that term $i$, $||pos_{target}-pos_{robot}||$ is the Euclidean distance from the arm to the target, $||a(t)||$ is the Euclidean norm of the action vector taken by the robot and $F_{target}$ is the current force applied to the target.
\begin{table}
\caption{Trained policies and their respective observation and action spaces}
\label{Table:params}

\resizebox{\columnwidth}{!}{%
\begin{tabular}{lll}
\hline
Policy & Observation Space & Action Space \\ \hline
\begin{tabular}[c]{@{}l@{}}Involuntary \\ Movement\end{tabular} & \begin{tabular}[c]{@{}l@{}}34 human joint values\\ 30 robot joint values\end{tabular} & \begin{tabular}[c]{@{}l@{}}10 human joint values\\ 7 robot joint values\end{tabular} \\ \hline
Weakness & \begin{tabular}[c]{@{}l@{}}34 human joint values\\ 30 robot joint values\end{tabular} & \begin{tabular}[c]{@{}l@{}}10 human joint values\\ 7 robot joint values\end{tabular} \\ \hline
\begin{tabular}[c]{@{}l@{}}Limit Range \\ of Motion\end{tabular} & \begin{tabular}[c]{@{}l@{}}34 human joint values\\ 30 robot joint values\end{tabular} & \begin{tabular}[c]{@{}l@{}}10 human joint values\\ 7 robot joint values\end{tabular} \\ \hline
blending network & \begin{tabular}[c]{@{}l@{}}Only System Parameters:\\ 1 for Estimate Weakness\\ 1 for Estimated Range Limit\\ 10 for Estimated \\ Involuntary \\ Movement Joints\end{tabular} & \begin{tabular}[c]{@{}l@{}}3 weighted values for \\ blending the policies\end{tabular} \\ \hline
\end{tabular}
}
\end{table}
\subsection{Training of Blending Network}
Similar to the sub-policies, we consider the same reward and utilize a PPO model with 2 layers of 64 nodes each for training the blending network. However, for the blending network we train for 400k time steps and the state space only consists of the system parameters. Unlike the sub-policies, our blending network is trained on a human with all three impairments and as such must consider many more cases of how the robot needs to act. By training the policy on a general all three impairments, we allow our blending network to become more robust to parameter identification and improve on the notion that training a single policy to handle all three impairments is complex and time-consuming due to sample inefficiency.

\subsection{Training of Domain Randomization}
To train the domain randomization model, we train on a human invoking exhibiting all three impairments. Similar to above, we use PPO with  layers of 64 nodes each. However, these impairments are now sampled uniformly as such:
\begin{enumerate}[label=(\alph*)]
  \item Involuntary Movement: The noise for each joints angle is between $[-10, 10]$ degrees.
  \item Weakened Strength: We consider a weakness coefficient between $[0.25, 1]$.
  \item Limited Range: We consider range limitations between $[.5, 1]$ times the original motion.
\end{enumerate}
\begin{table}[]
 \caption{Trained policies and their respective observation and action spaces}
\label{Table:methods}

\resizebox{\columnwidth}{!}{%
\begin{tabular}{lll}
\hline
Method & Policy Blending & State Space \\ \hline
\begin{tabular}[c]{@{}l@{}}Domain \\ Randomization\end{tabular} & No & \begin{tabular}[c]{@{}l@{}}Trained from Human \\ and Robot Observation \\ Space\end{tabular} \\ \hline
UKF & Yes & \begin{tabular}[c]{@{}l@{}}12 Parameters Estimated \\ By UKF Sampling Real World\end{tabular} \\ \hline
Autotuned SPM & Yes & \begin{tabular}[c]{@{}l@{}}12 Parameters Estimated by \\ Mapping Function of Interaction \\ Between Policy and Real World\end{tabular} \\ \hline
\begin{tabular}[c]{@{}l@{}}Perfect \\ Parameters\end{tabular} & Yes & \begin{tabular}[c]{@{}l@{}}Parameters are Passed as the \\ State Space at the Start of \\ Each Epsiode\end{tabular} \\ \hline
\end{tabular}
}
\end{table}

\section{Discussion and Results}

\begin{figure*}[!htb]
\minipage{0.2\textwidth}
  \includegraphics[width=\linewidth]{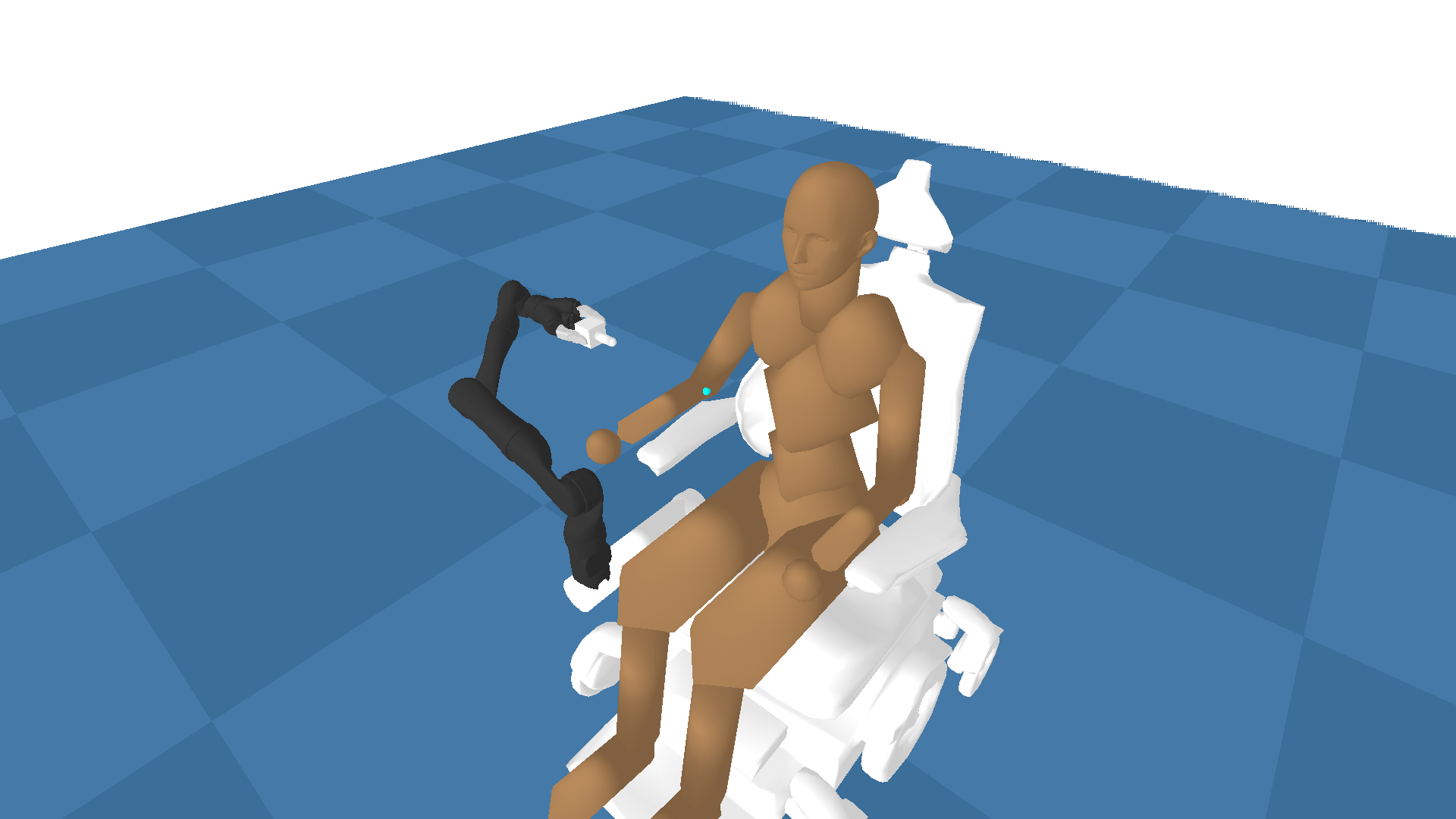}
\endminipage\hfill
\minipage{0.2\textwidth}
  \includegraphics[width=\linewidth]{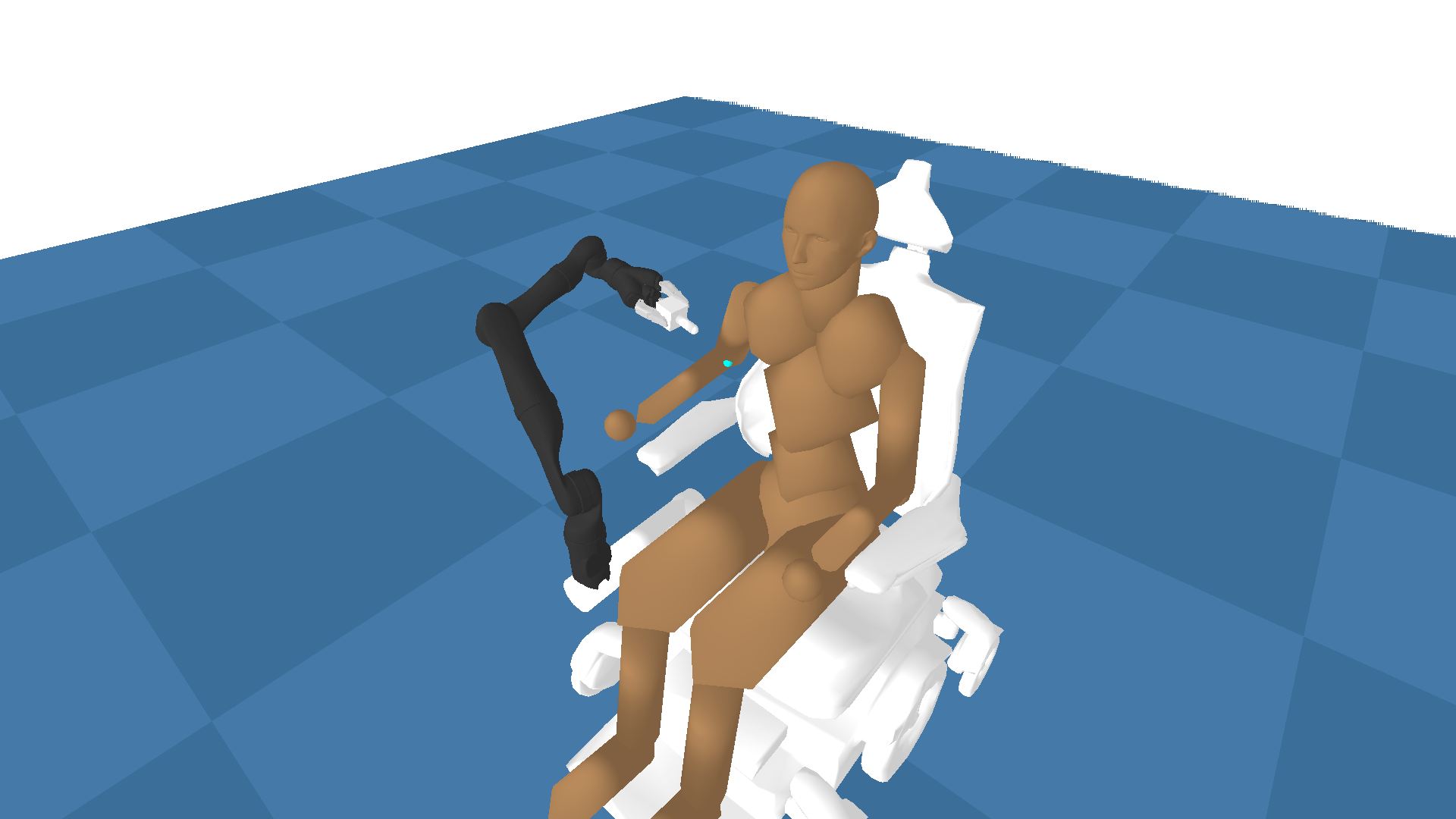}
\endminipage\hfill
\minipage{0.2\textwidth}%
  \includegraphics[width=\linewidth]{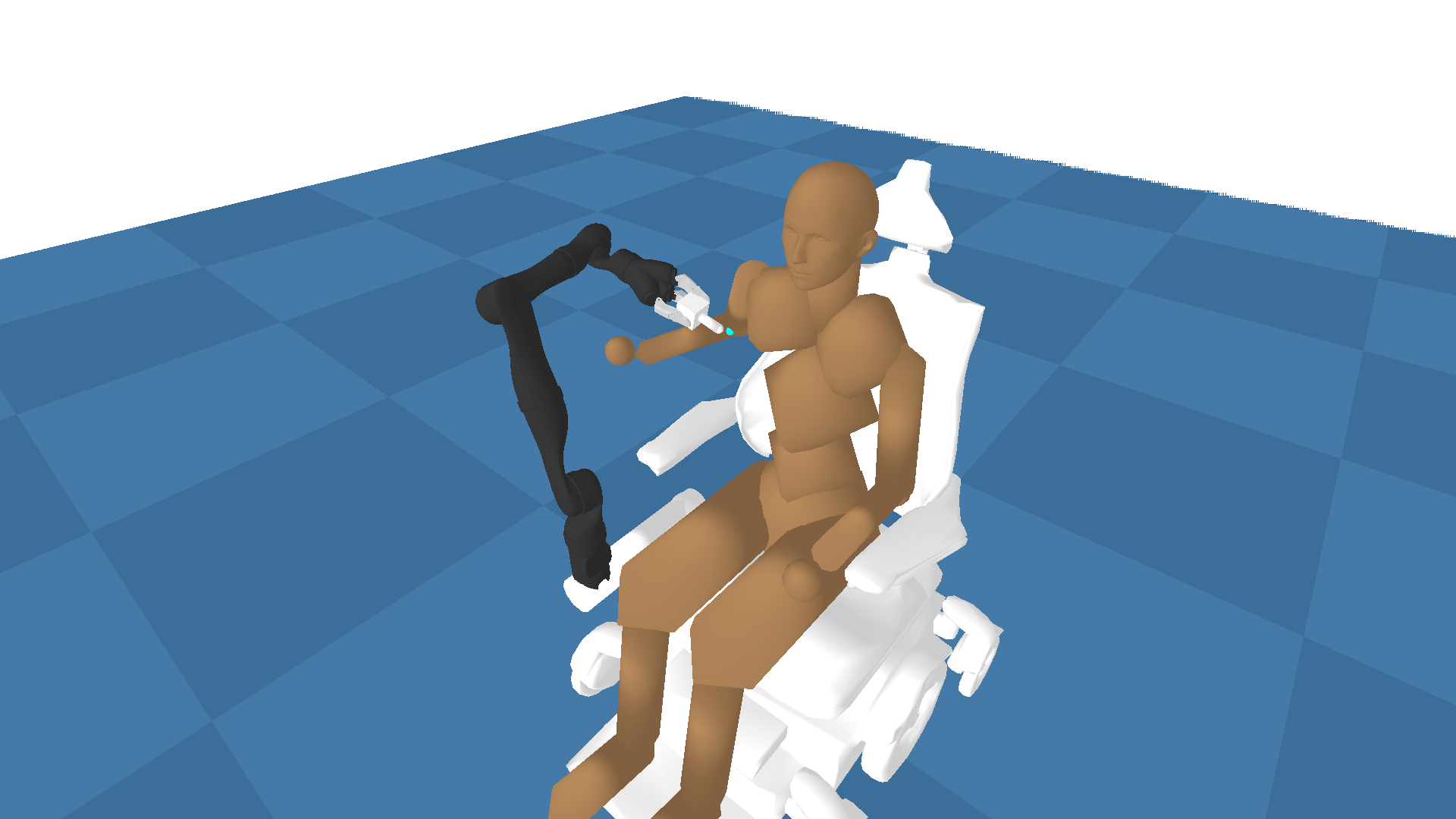}
\endminipage
\minipage{0.2\textwidth}%
  \includegraphics[width=\linewidth]{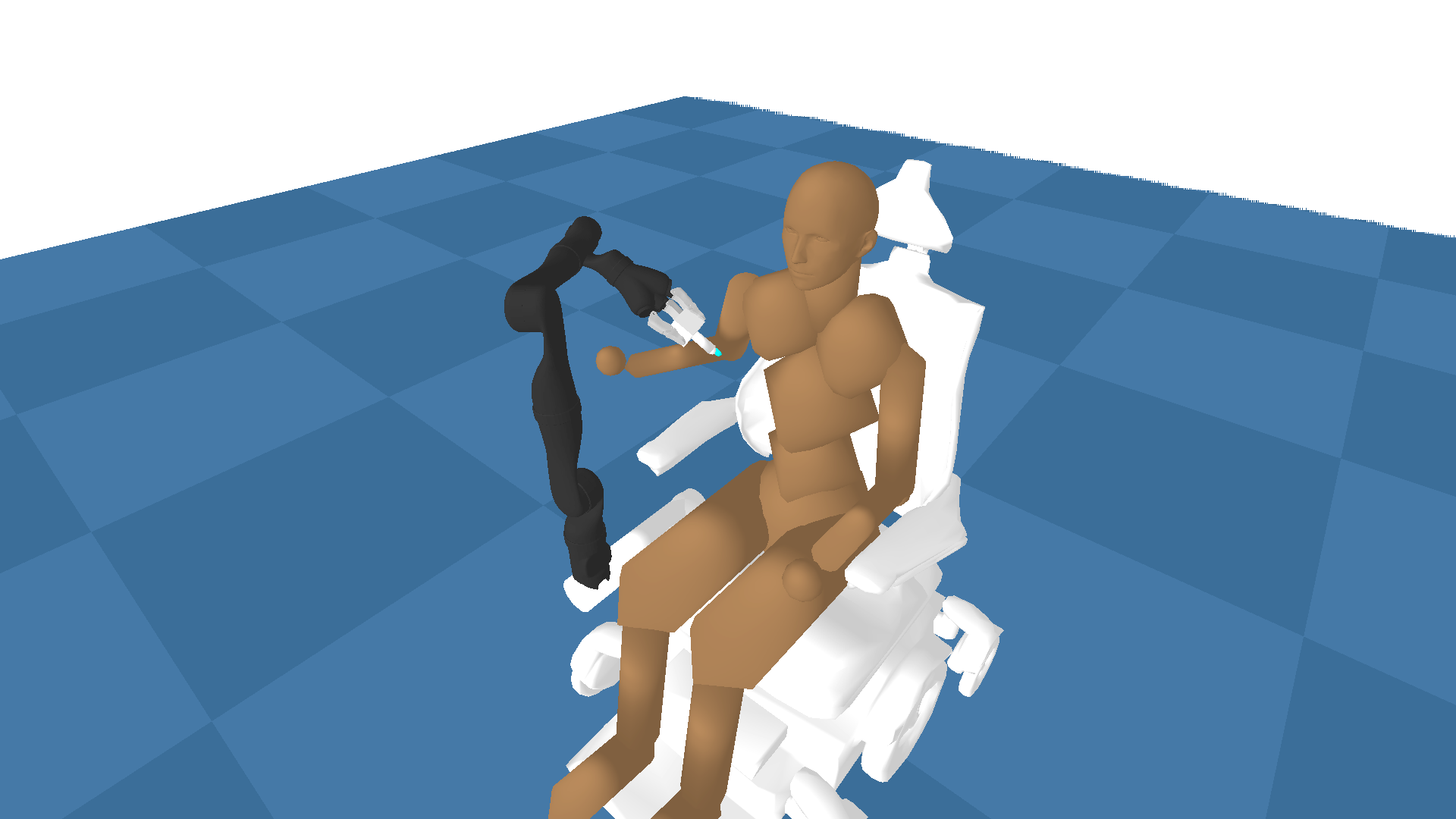}
\endminipage
\minipage{0.2\textwidth}%
  \includegraphics[width=\linewidth]{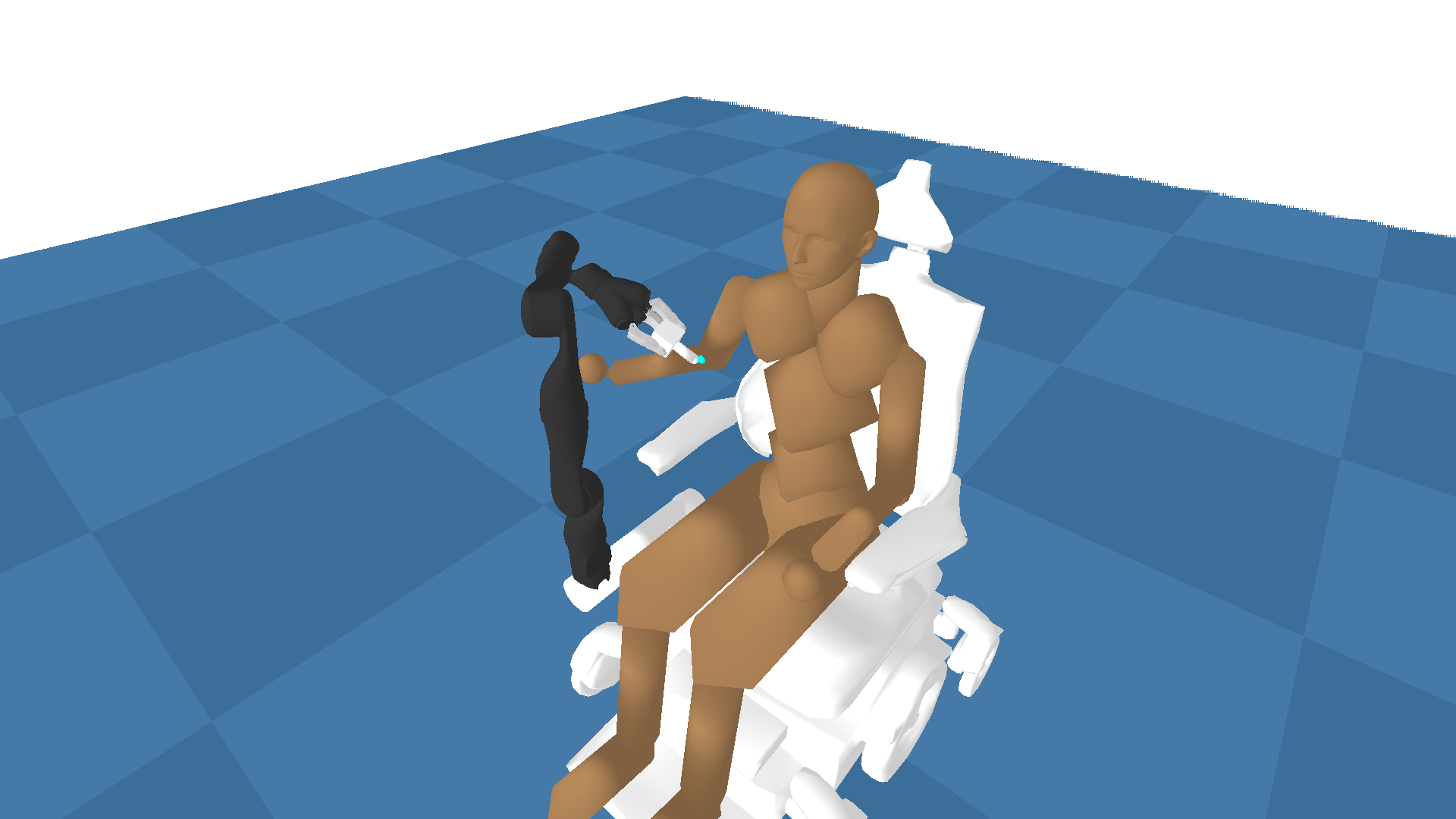}
\endminipage
\caption{Example of our robot completing the itching task even when the human is dis-functionally moving its arm upward}
\label{fig:scratch}
\end{figure*}
\begin{figure}[h!]
  \includegraphics[width=0.5\textwidth]{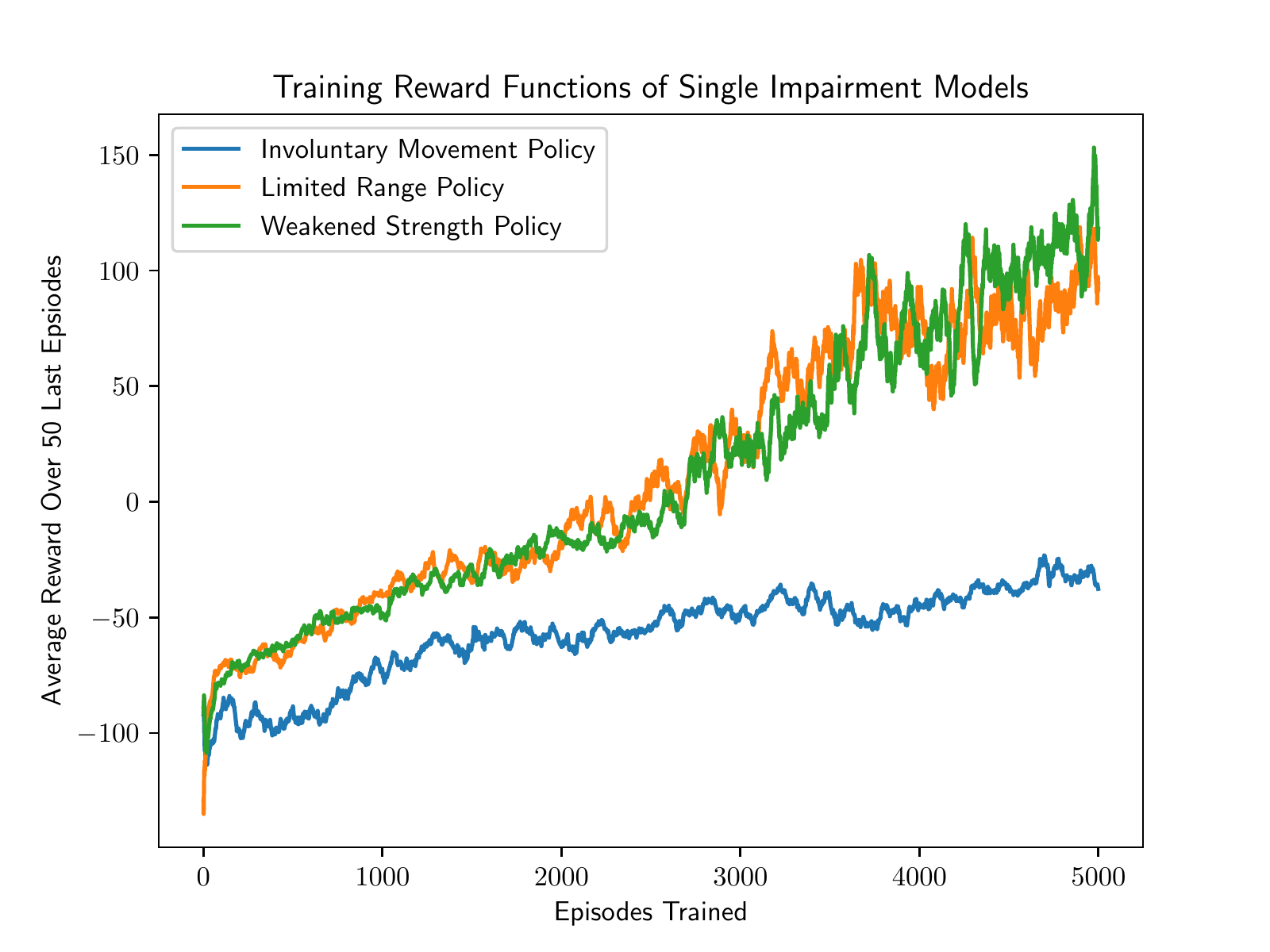}
  \caption{Training reward average over 50 episodes for single impairment policies. The rewards here are averaged over the training of 3 seeds.}
  \label{fig:rewardSingle}
\end{figure}
\begin{figure}
\begin{minipage}{0.95\textwidth}
\includegraphics[width=0.24\textwidth]{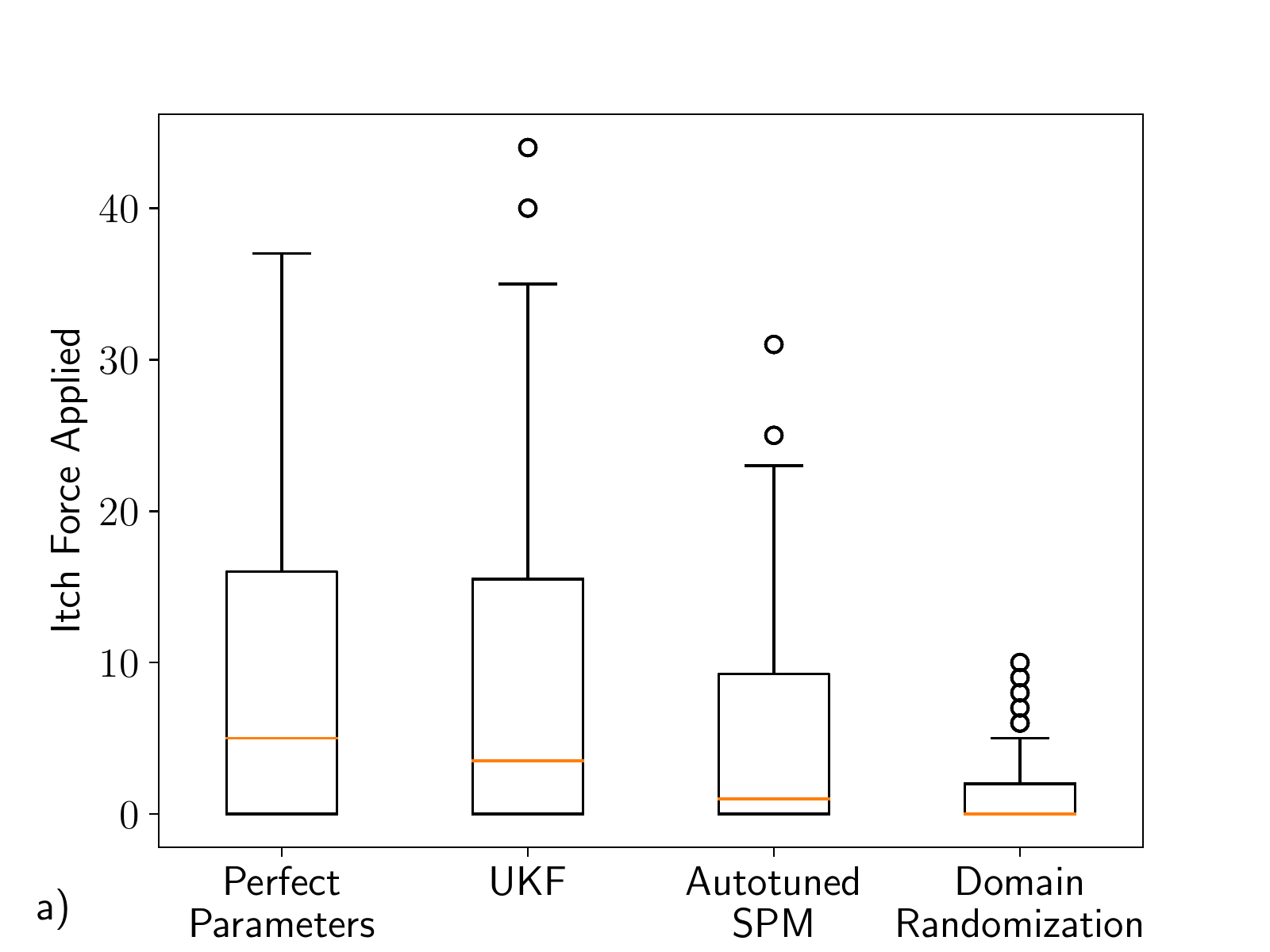}
\includegraphics[width=0.24\textwidth]{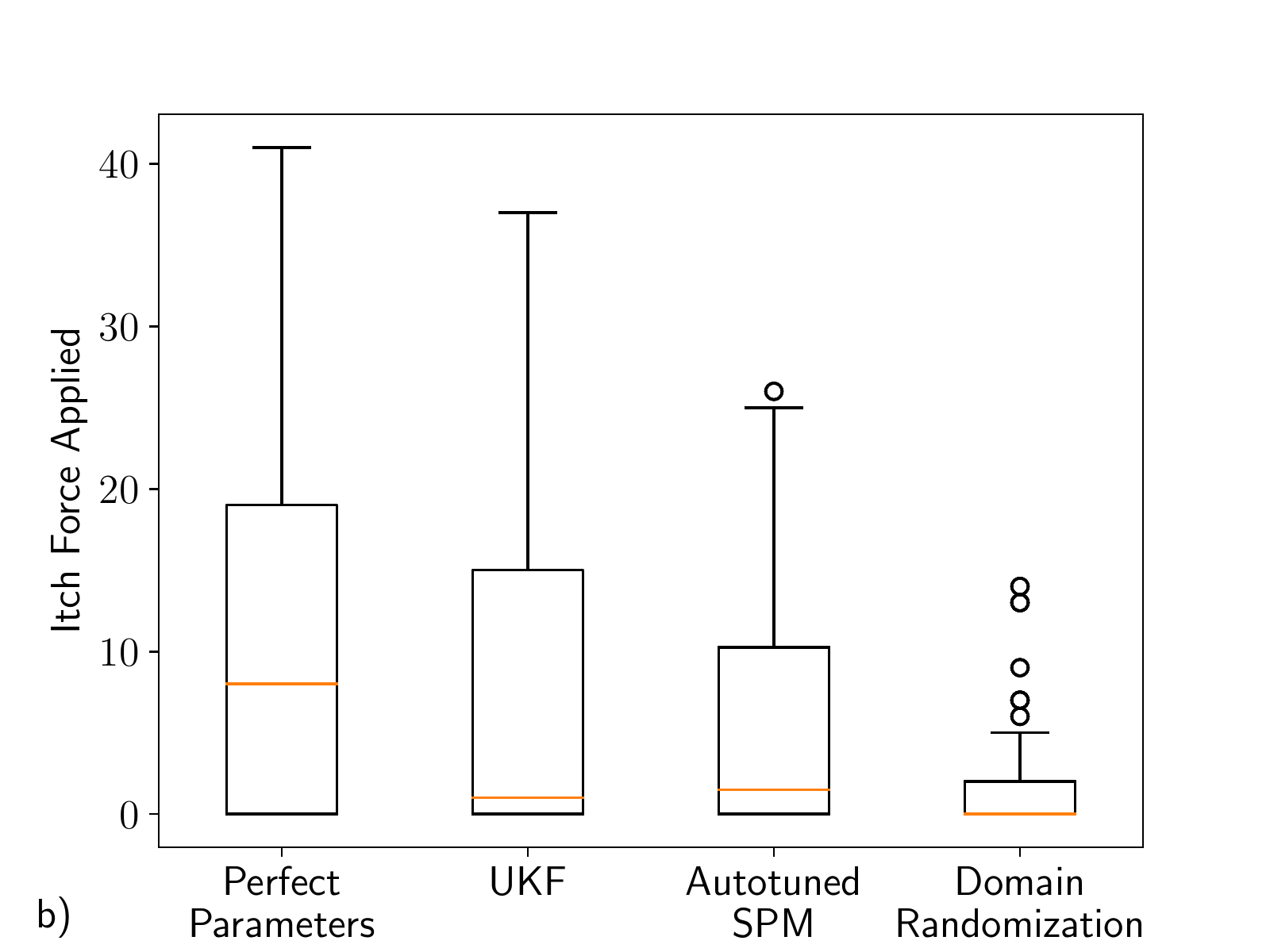}\\\quad
\includegraphics[width=0.24\textwidth]{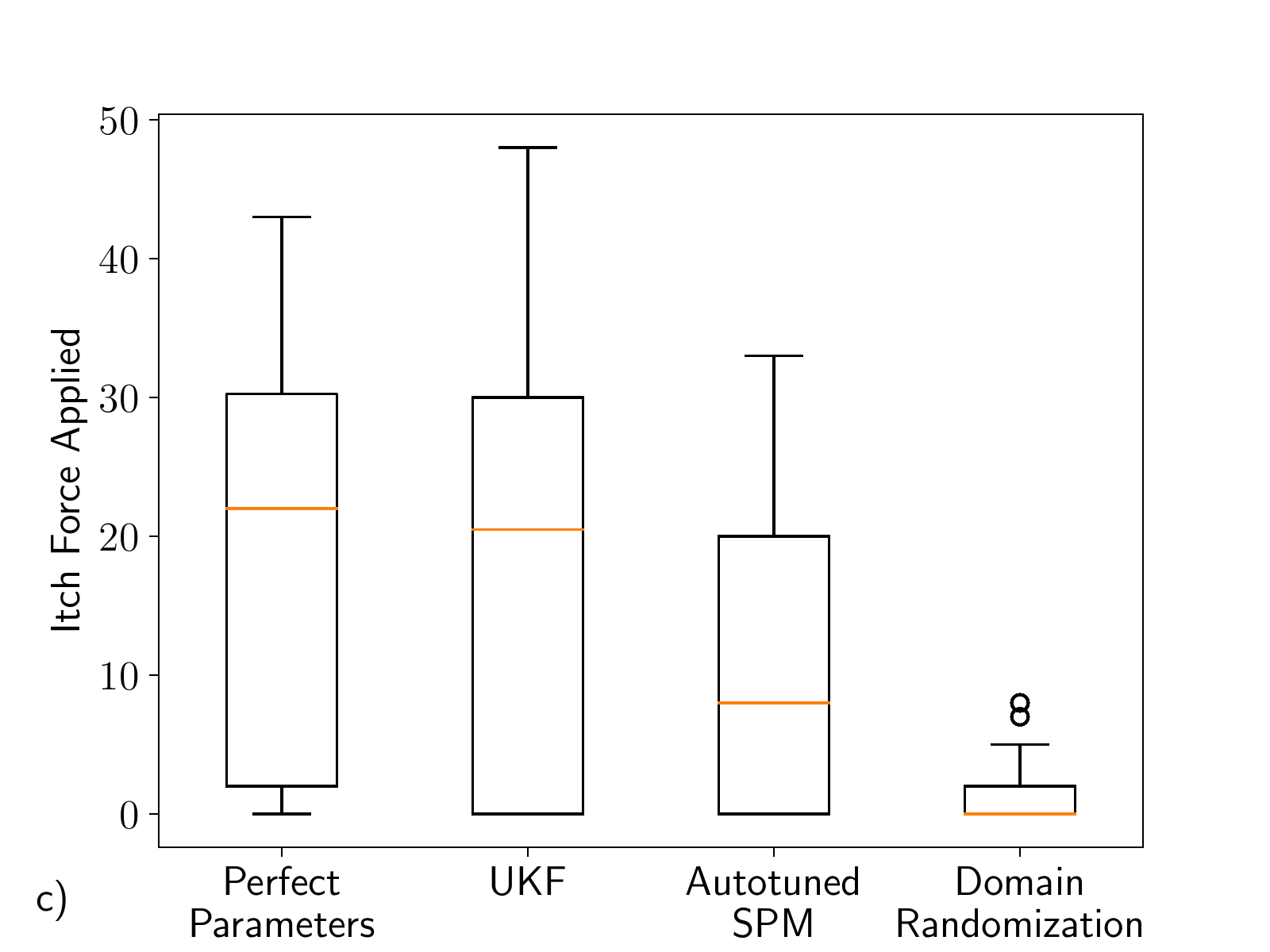}
\includegraphics[width=0.24\textwidth]{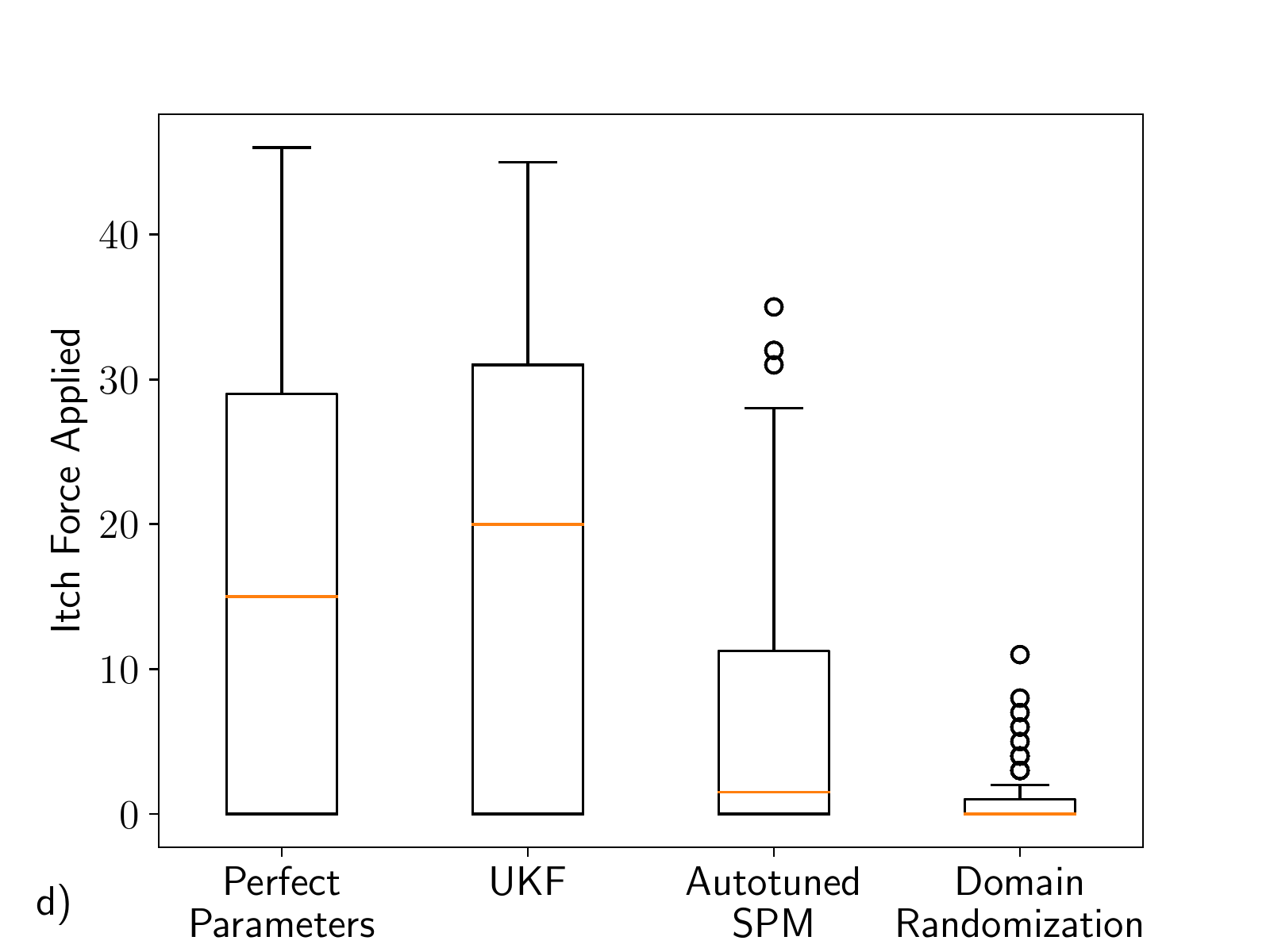}
\end{minipage}
\caption{Application of the trained policy to a real environment for 100 separate episodes. We use the highest reward policy for each situation. a) Shows the itch force applied when the human has a combination of all three impairments. b), c), and d) show the force applied when the human has a single impairment in the form of limited range, weakness, or involuntary motion respectively.}
\label{fig:result}
\end{figure}
\begin{figure}[h!]
  \includegraphics[width=0.45\textwidth]{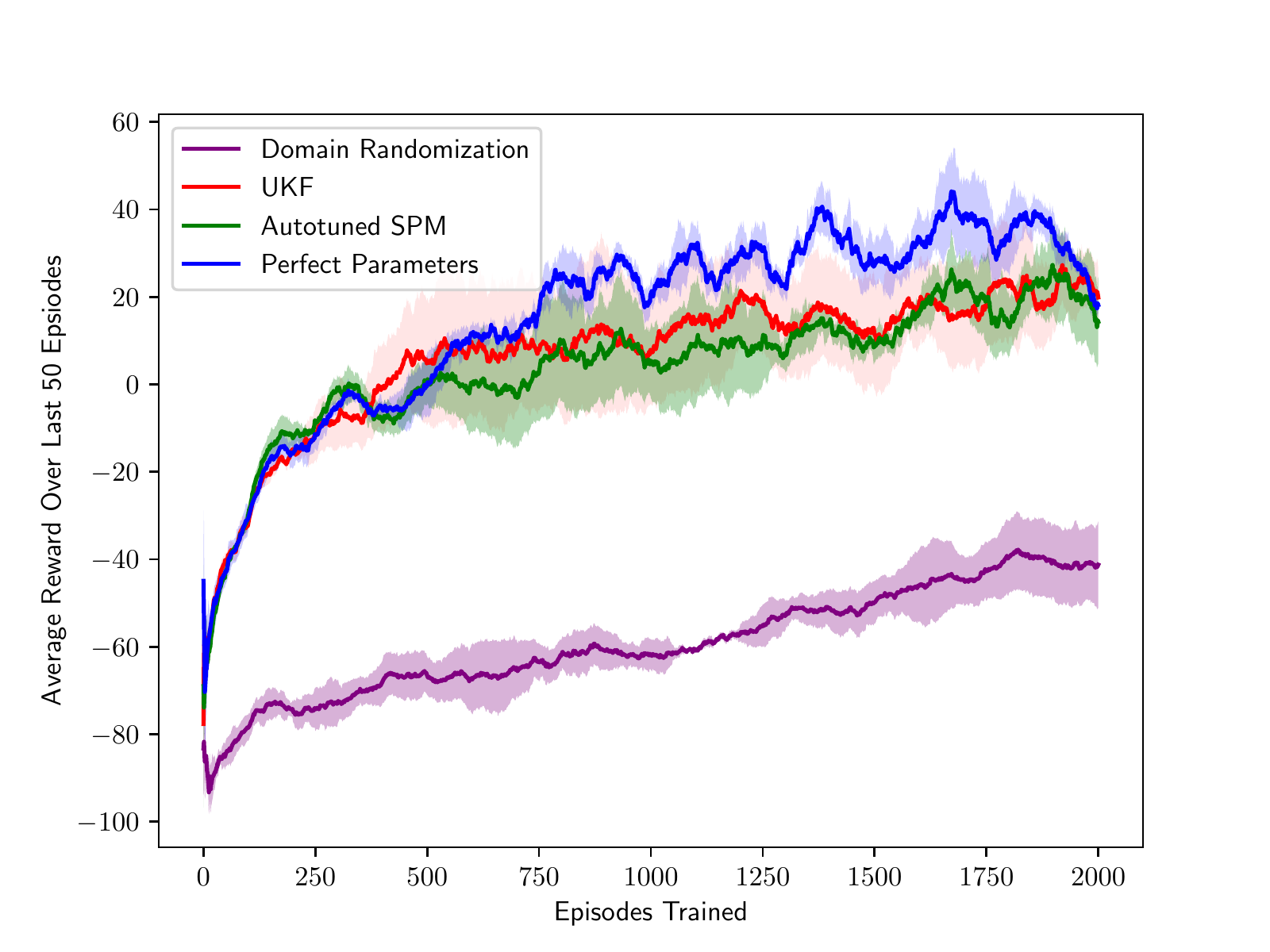}
  \caption{Training reward averaged over 50 episodes for our policy exploration methods. The rewards shown here are averaged over the training of 3 seeds}
  \label{fig:reward}
\end{figure}

For our initial sub-policies, we can see that the weakness and limit based policies can achieve a higher reward consistently over the involuntary movement policy in Fig. \ref{fig:rewardSingle}. For our blending network, we consider the best performing sub-policies and only train on humans with a combination of all three impairments. As such, in Fig. \ref{fig:reward} can see that the rewards are much lower than those of the individual policies. Additionally, we notice that there is a significant advantage to using a blending-based policy with system identification over general domain randomization. Furthermore, we can see that the ability to estimate the real-world parameters enhances the policies overall convergence as the auto tuned policies struggle to achieve the same success as the UKF based or the baseline (system parameters are perfectly known to the blending network at each timestep). 
To further evaluate our policies, we define a testing experiment in which we undergo 100 episodes of our human exhibiting all three impairments in which the impairment values are sampled as above but in conjunction. We still utilize the given system identification method for estimating the state space of the blending network.  Fig. \ref{fig:result} a) shows a box plot of the performance for the joint parameter variations. We consider experiments in which the human only enacts a single impairment and the results are shown in b), c), and d) of Fig. \ref{fig:result}  and Table \ref{Table:stats}, respectively. 

From the box plots, we can see that we outperform domain randomization for 100 separate episodes. Furthermore, there is a difference between the system identification methods as the UKF and the system fed with the correct parameters outperform the autotuned search approach. As such, we can determine two important things about our approach. First, the policy blending has a significant improvement over general domain randomization in terms of both sample efficiency and performance. Second, our design can successfully employ various types of system identification; however, those identification methods may significantly affect the overall performance of the policy and should be based on the maximum amount of domain knowledge available. 

 Given this, we must note limitation of our scheme is that we need to develop the sub-policies; however, these theoretically provide us stability and robustness when faced with unknown environments.  Additionally, given that these sub-policies can be reused as they are now decoupled from the main blending network, different approaches can quickly be tested and tuned - a problem limiting current domain randomization methods. Furthermore, it is not guaranteed that a linear combination of weights from the blending network and the action spaces of the blending policy is a good approximation other than from an empirical standpoint. Different methods could certainly be used for the policy blending and this is a future direction of exploration. 

\begin{table}[]
  \caption{Mean and STDEV of Each Method Given a Specific Impairment } 
  \label{Table:stats}
\resizebox{\columnwidth}{!}{%
\begin{tabular}{lllllllll}
\hline
 & \multicolumn{2}{l}{\begin{tabular}[c]{@{}l@{}}Combined\\ Impairments\end{tabular}} & \multicolumn{2}{l}{\begin{tabular}[c]{@{}l@{}}Involuntary\\ Movement\\ Impairment\end{tabular}} & \multicolumn{2}{l}{\begin{tabular}[c]{@{}l@{}}Limited \\ Range \\ of Motion\end{tabular}} & \multicolumn{2}{l}{\begin{tabular}[c]{@{}l@{}}Weakness \\ in Joints\end{tabular}} \\ \hline
Method & Mean & STDEV & Mean & STDEV & Mean & STDEV & Mean & STDEV \\ \hline
\begin{tabular}[c]{@{}l@{}}Domain \\ Randomization\end{tabular} & 1.33 & 2.18 & 0.96 & 1.94 & 1.56 & 2.77 & 1.07 & 1.90 \\ \hline
UKF & 8.68 & \textbf{10.58} & \textbf{18.14} & \textbf{14.62} & 8.05 & 10.09 & 18.13 & \textbf{14.49} \\ \hline
Autotuned SPM & 5.26 & 7.35 & 6.34 & 8.67 & 5.71 & 7.35 & 10.42 & 10.32 \\ \hline
\begin{tabular}[c]{@{}l@{}}Perfect \\ Parameters\end{tabular} & \textbf{9.03} & 10.23 & 15.02 & 14.40 & \textbf{11.2} & \textbf{11.71} & \textbf{19.0} & 13.81 \\ \hline
\end{tabular}
}
\end{table}

\section{Conclusions and Future Work}
In this work, we present a concurrent policy blending and system identification scheme for learning a generalized policy with respect to varying system parameters. With this scheme, we demonstrate the ability to solve a collaborative human and robot task in which the human is impaired with multiple separate, but impactful conditions. Additionally, we demonstrate that our policy outperforms the sample inefficient domain randomization as we can utilize diverse system identification methods to significantly improve over a single general policy. As such, in this work, we provide a framework for efficiently training generalized policies that are robust to an ever changing system.

\bibliographystyle{IEEEtran}
\bibliography{bibfile}

\end{document}